\documentclass{article}

\usepackage{microtype}
\usepackage{graphicx}
\usepackage{subfigure}
\usepackage{booktabs} 
\usepackage{url}
\usepackage{hyperref}

\usepackage[accepted]{icml2024}

\usepackage{amsmath}
\usepackage{amssymb}
\usepackage{mathtools}
\usepackage{amsthm}

\usepackage[capitalize,noabbrev]{cleveref}

\theoremstyle{plain}

\theoremstyle{definition}

\theoremstyle{remark}

\usepackage[textsize=tiny]{todonotes}

\icmltitlerunning{CLLMs: Consistency Large Language Models}

\usepackage{multirow}

\usepackage{amsmath,amsfonts,bm}

\def\eqref#1{equation~\ref{#1}}

\def\1{\bm{1}}

\def\vl{{\bm{l}}}

\def\vx{{\bm{x}}}
\def\vy{{\bm{y}}}
\def\vz{{\bm{z}}}

\DeclareMathAlphabet{\mathsfit}{\encodingdefault}{\sfdefault}{m}{sl}
\SetMathAlphabet{\mathsfit}{bold}{\encodingdefault}{\sfdefault}{bx}{n}

\newcommand{\E}{\mathbb{E}}

\begin{document}

\twocolumn[
\icmltitle{CLLMs: Consistency Large Language Models}



\icmlsetsymbol{equal}{*}

\begin{icmlauthorlist}
\icmlauthor{Siqi Kou}{equal,sjtu}
\icmlauthor{Lanxiang Hu}{equal,ucsd}
\icmlauthor{Zhezhi He}{sjtu_ee}
\icmlauthor{Zhijie Deng}{sjtu}
\icmlauthor{Hao Zhang}{ucsd}

\end{icmlauthorlist}

\icmlaffiliation{sjtu}{Qing Yuan Research Institute, SEIEE, Shanghai Jiao Tong University}
\icmlaffiliation{sjtu_ee}{School of Electronic Information and Electrical Engineering, Shanghai Jiao Tong University}
\icmlaffiliation{ucsd}{University of California, San Diego}

\icmlcorrespondingauthor{Zhijie Deng}{zhijied@sjtu.edu.cn}

\icmlkeywords{Machine Learning, ICML}

\vskip 0.3in
]



\printAffiliationsAndNotice{\icmlEqualContribution} 

\begin{abstract}
Parallel decoding methods such as Jacobi decoding show promise for more efficient LLM inference as it breaks the sequential nature of the LLM decoding process and transforms it into parallelizable computation. However, in practice, it achieves little speedup compared to traditional autoregressive (AR) decoding, primarily because Jacobi decoding seldom accurately predicts more than one token in a single fixed-point iteration step. To address this, we develop a new approach aimed at realizing fast convergence from any state to the fixed point on a Jacobi trajectory. This is accomplished by refining the target LLM to consistently predict the fixed point given any state as input.
Extensive experiments demonstrate the effectiveness of our method, showing 2.4$\times$ to 3.4$\times$ improvements in generation speed while preserving generation quality across both domain-specific and open-domain benchmarks. Our code is available at \href{https://github.com/hao-ai-lab/Consistency_LLM}{https://github.com/hao-ai-lab/Consistency\_LLM}.






\end{abstract}



\section{Introduction}
\label{intro}


Large language models (LLMs), including GPT-4~\citep{achiam2023gpt}, LLaMA~\citep{touvron2023llama,touvron2023llama2}, 
PaLM~\citep{anil2023palm}, are pushing the limit of artificial intelligence. As LLMs are integrated into more applications~\cite{zheng2023judging, wu2023next}, 
the inference latency of LLMs plays a crucial role in ensuring a positive user experience and high service quality. 
However, LLM serving operates in an AR paradigm, generating one token at a time due to the attention mechanism's need for previous token states to generate the next one. To produce a lengthy response, one must execute forward passes through the LLMs as many times as the number of tokens generated, resulting in high latency.



Existing methods address this issue from various perspectives.
For example, speculative decoding~\cite{leviathan2023fast,chen2023accelerating} introduces a small draft LLM to guess tokens and let the target LLM verify them in parallel. 
Although they can opportunistically generate multiple tokens in a single evaluation of the target LLM, obtaining a small yet effective draft model is non-trivial; managing multiple models within a single system remains a challenging engineering task. 
Medusa~\cite{cai2024medusa} alternatively augments the target LLM with extra guess heads to enable self-speculation with as much as $3\times$ speedup on various tasks. 
Yet, the number of added parameters can be significant (e.g., Medusa2 with 5 extra heads adds 1.6B parameters for a 6.7B target LLM). Increased memory consumption could limit generation length and negatively affect inference latency due to the reduction in available memory for key-value (KV) cache~\citep{pope2023efficiently}. 



\begin{figure}[t]
\centering
\includegraphics[width=0.49\textwidth]{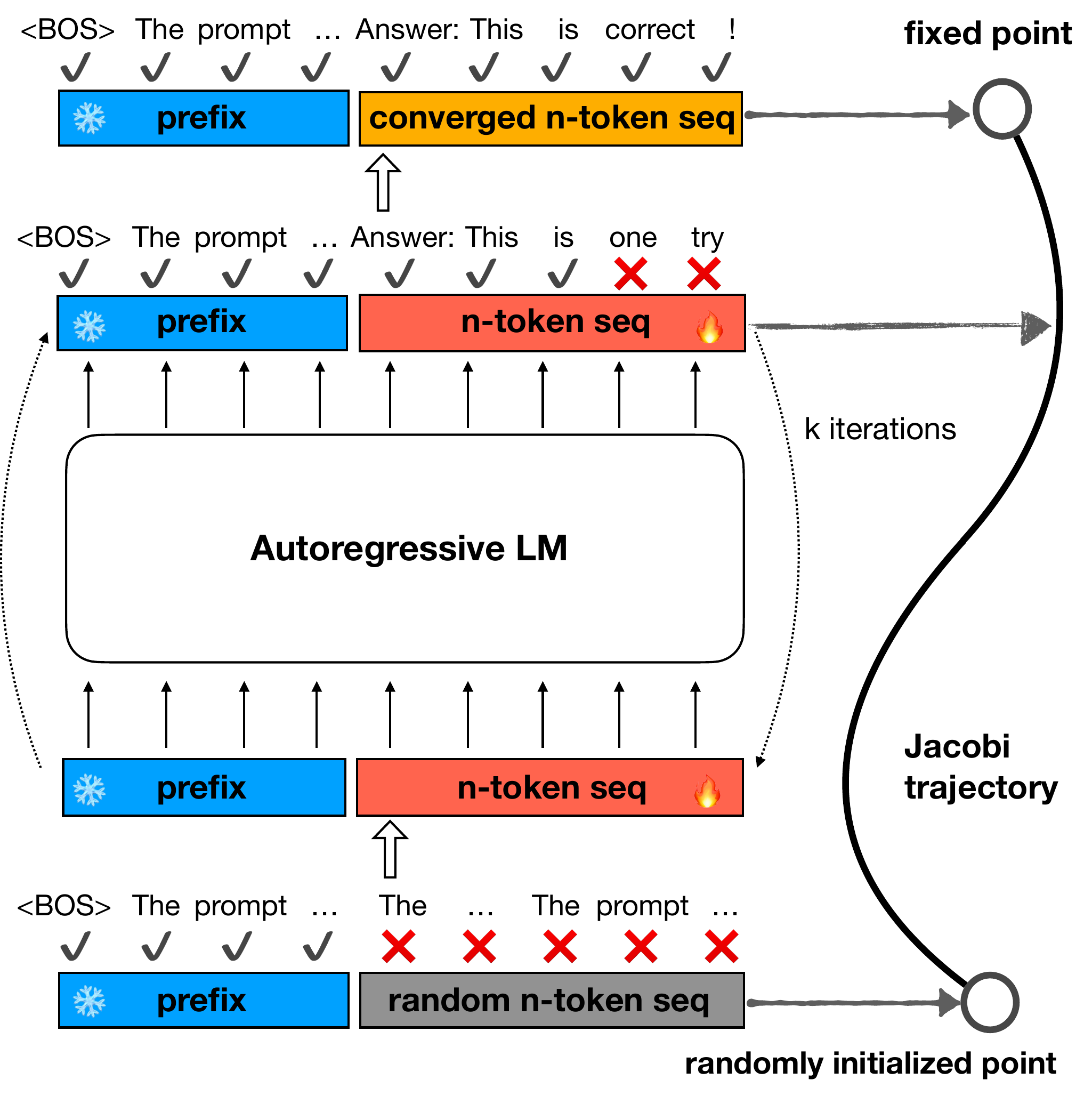} 
\vskip -0.1in
\caption{An instance of Jacobi trajectory. ``$n$-token seq'' refers to the $n$-token sequence that is iteratively updated in Jacobi iterations.}
\label{fig:jacobi_visualization}
\vskip -0.2in
\end{figure}

On the other hand, originating from the Jacobi and Gauss-Seidel fixed-point iteration for solving nonlinear equations~\citep{ortega2000iterative,song2021accelerating}, the Jacobi decoding method ~\cite{santilli2023accelerating} first randomly guesses the next $n$ tokens in a sequence (referred to as $n$-token sequence hereinafter) from an input prompt. The $n$-token sequence, along with the prompt, is then fed to the LLM to iteratively update itself. Eventually, the $n$-token sequence converges to the same output generated by AR decoding under a greedy strategy (see Figure~\ref{fig:jacobi_visualization}). The evolution of the $n$-token sequence forms a \textbf{Jacobi trajectory} between a randomly initialized sequence to the $n$-token sequence generated by AR decoding (i.e., the fixed point). 




However, vanilla Jacobi decoding for LLMs shows only marginal speedup over AR decoding in practice, e.g., an average of 1.05$\times$ speedup in \citet{santilli2023accelerating}. 
This is because a LLM can rarely yield a correct token when there are incorrection\footnote{By correctness, we mean alignment with the AR decoding result under a greedy sampling strategy.} in its preceding tokens due to the attention mechanism, resulting in a long trajectory as illustrated on the left side of Figure~\ref{fig:trajectory_compare}. Lookahead decoding~\citep{fu2024break} improves the efficiency by leveraging n-grams generated from previous Jacobi iterations and verify them in parallel during the decoding process. However, both work are unable to achieve the same level of speedup as Meudsa.

This work aims to achieve all three goals by refining the target LLM. 
Specifically, we propose to fine-tune the LLM so that it can yield multiple, instead of one, subsequent tokens of a prefix at once. 
In the ideal case, with the prompt and a randomly initialized $n$-token sequence as input, our goal is to train a LLM that can generate the same $n$-token sequence as AR decoding (the fixed point) using only one step.
Our preliminary experiments show the single-step learning task is difficult when $n$ is large, and leads to slow model convergence. We therefore ease the learning process by also taking intermediate points on the Jacobi trajectory with more correct tokens into account. 
In particular, for the second to last point on the trajectory, the learning is identical to AR modeling, at which the target LLM without adaptation has already excelled. 




We argue such a learning strategy that a single model is tuned to solve a series of learning problems of mapping any arbitrary point on the trajectory to the fixed-point is beneficial to model convergence (see ~\cref{fig:objective_illustration_global} and ~\cref{fig:objective_illustration_local}).
Imagining the evolution of the $n$-token sequence as the denoising process of a natural image~\cite{ho2020denoising,song2021scorebased}, we surprisingly find that the above learning procedure draws a sharp analogy to the acceleration technique for diffusion models named consistency models (CMs)~\cite{songconsistency,song2023improved}. CMs aim to achieve single-step image generation using the denoising objective by minimizing distances between consecutive denoising steps along the probability flow ordinary differential equation (ODE) trajectory during training. 
Our method and CMs share the notion of directly mapping intermediate states of a solving process (of non-linear systems or ODEs) to its final solution for inference acceleration. 
Based on these, we refer to our trained models as Consistency Large Language Models (CLLMs). In comparison with previous methods like speculative decoding and Medusa, CLLM doesn't introduce extra memory cost to accommodate auxiliary model components while delivering significant speedup with minimal performance degradation.

To implement this learning strategy, it only requires model training with two loss terms.
Following CMs, we can convert the aforementioned learning objective into a consistency loss where the model is demended to map arbitrary point on the Jacobi trajectory to the fixed point. 
CLLMs also include an AR loss to avoid deviating from the distribution of the target LLM and hence ensure the generation quality. 

The fine-tuning cost of CLLMs is moderate, e.g., training on only $\sim1$M tokens for LLaMA-7B to achieve a $3.4\times$ speedup on the Spider dataset. 
We further empirically identify that such acceleration is likely to stem from the existence of 1) \emph{fast forwarding}, where multiple consecutive tokens are correctly predicted in a single forward pass, and 2) \emph{stationary tokens}, which are correctly predicted and remain unaltered through subsequent iterations, despite being preceded by inaccurate tokens. An illustration of the examples is shown in Figure~\ref{fig:trajectory_compare}. 

To summarize, our key contributions are as follows:
\begin{list}{$\bullet$}{\leftmargin=1em \itemindent=0em}
    \item We propose Consistency Large Language Models (CLLMs), a new family of LLMs specialized for the Jacobi decoding method for latency reduction. 
    \item We empirically observe the existence of \emph{fast forwarding} and \emph{stationary tokens} phenomena in Jacobi decoding of CLLMs. Empirically, CLLMs can lead to a 2.0$\times$ to 6.8$\times$ improvement in the count of fast-forwarded tokens and stationary tokens compared to the original LLM. 
    \item We demonstrate the efficacy of CLLMs on a variety of benchmarks. On domain-specific benchmarks including GSM8K, CodeSearchNet Python, and Spider, CLLMs can achieve 2.4$\times$ to 3.4$\times$ speedup using Jacobi decoding with nearly no loss in accuracy. 
    On open-domain benchmark MT-bench, CLLMs can achieve 2.4$\times$ speedup on ShareGPT with state-of-the-art performance, scoring 6.4. 
\end{list}

\section{Related Work}

\textbf{Efficient LLM Inference.} This body of work can be broadly categorized into two streams: methods that necessitate additional training and those that do not. 
The high AR inference cost in LLMs has sparked a surge in research aimed at efficient LLM inference, primarily focused on accelerating the AR decoding process. 


The methods that do not require additional training include speculative decoding, as introduced in studies by \citet{leviathan2023fast} and \citet{chen2023accelerating}. These techniques enhance LLM decoding speed by leveraging a smaller draft model to predict the outputs of a larger target model which subsequently verifies these predictions. Another category of training-free approaches involves system- or hardware-oriented optimizations. Notable examples include PagedAttention~\citep{kwon2023efficient}, which optimizes KV cache management for throughput using memory paging, and FlashAttention~\citep{dao2022flashattention, dao2023flashattention}, which accelerates attention module computations by reducing HBM access via softmax tiling. Other strategies enhance LLM inference speed by optimizing model designs, reducing weight/activation precision, and utilizing sparsity, including multi-query and grouped-query attention mechanisms with fused heads~\citep{shazeer2019fast, ainslie2023gqa}, post-training quantization~\citep{dettmers2022llm, xiao2023smoothquant, frantar2022gptq, lin2023awq}, and various pruning techniques~\citep{sun2023simple, frantar2023sparsegpt, ashkboos2024slicegpt}.

For methods that necessitate training, they often require integration of auxiliary components, such as additional LM or AR heads, to facilitate faster AR generation~\citep{cai2024medusa, li2024eagle}. It may also involve significant modifications to the model weights or architecture, as seen in various pruning approaches~\citep{ma2023llm, xia2022structured, xia2023sheared}. Moreover, training can enhance certain training-free techniques, like speculative decoding, by capturing the behavior of the original, larger model in a smaller student model through distillation, thereby retaining performance with reduced size~\citep{zhou2023distillspec, liu2023online}. 
An detailed analysis that compare CLLMs with different SOTA baseline methods are further discussed and compared in Section~\ref{appendix:baseline_comparison} and Table~\ref{tab:baseline_comparison}. It's worthy noticing that CLLMs requires neither modification to pre-trained models nor any auxiliary components. This brings higher memory efficiency and adaptability to users at inference time. 


\textbf{LLM Distillation.} Knowledge distillation (KD) serves as a technique for creating smaller models that replicate the functionality of larger ones. While traditional KD approaches often fall short for LLMs, \citep{gu2023knowledge} has adapted KD for autoregressive LLMs, focusing on minimizing the reverse KL divergence between student and teacher models through student-driven decoding. In another advancement, \citet{agarwal2023gkd} introduces generalized knowledge distillation (GKD), which balances forward and reverse KL divergences by employing a mix of data sampled from both teacher and student models.

CLLMs are distinct from these works as our proposed method can be regarded as a self-distillation approach with a Jacobi trajectory training dataset that matches the target LLM’s output distribution.

\textbf{Consistency Models.} 
Diffusion models~\citep{ho2020denoising, song2021scorebased} suffer from slow iterative sampling process. Consistency models overcome this limitation by mapping any point along the probability flow ODE of the diffusion process back to the original point, corresponding to the initial image, in a single step~\citep{songconsistency}. In this work, we highlight that a parallelism can be drawn between the few-step generation capability of CLLMs and that of the consistency models.

\section{Methodology}
This section begins with a review of the Jacobi decoding method~\citep{santilli2023accelerating} for accelerating LLM inference, then elaborates on CLLMs, a refinement of pre-trained LLMs to enjoy higher speedup from Jacobi decoding. In this paper, we only consider greedy sampling and leave other sampling strategies to future work. We also empirically identify the \emph{fast-forwarding} phenomenon and the emeragence of \emph{stationary tokens} from CLLMs, which serve as the source of such acceleration. 

\subsection{Preliminary: Jacobi Decoding}
\label{sec:preliminary}

\begin{figure*}[thb]
\centering
\includegraphics[width=0.99\textwidth]
{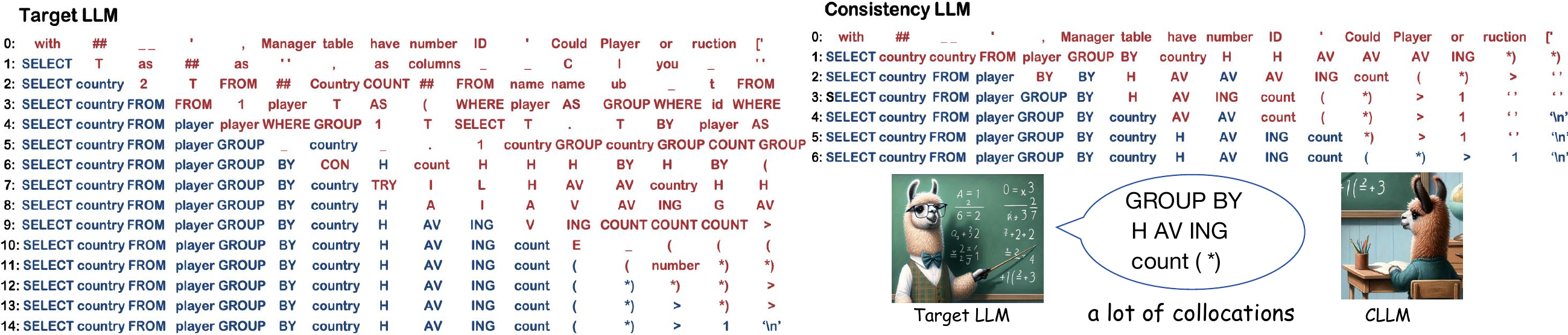} 
\caption{Comparison of Jacobi trajectory between a target LLM and CLLMs on Spider. Each point along the Jacobi trajectory is a color-coded sequence: blue for correct tokens matching with AR results, and red for inaccurate ones. CLLM demonstrates enhanced efficiency, converging to the fixed point $2\times$ faster than the target LLM. This increased efficiency in the CLLM can be attributed to the consistency loss which facilitates the learning of the structure of each $n$-token sequence given a prefix.
}
\label{fig:trajectory_compare}
\vskip -0.1in
\end{figure*}

Given a prompt $\vx$ and a pre-trained LLM $p(\cdot| \vx)$, we obtain the model response typically with the standard AR decoding method under the greedy strategy, \textit{i.e.},
\begin{equation}
\label{eq:ar_decoding}
\begin{aligned}
y_i = \underset{y}{\mathrm{arg\,max}}\ p(y | \vy_{<i}, \vx) \;\, \text{for}\,\, i = 1,\dots,n
\end{aligned}
\end{equation}
where $\vy_{<i}$ denotes $\{y_{1},  \ldots, y_{i-1} \}$.
As shown, $n$ forward passes of the LLM are required to obtain $n$ tokens $\vy_{\leq n}$. 
The sequential nature of AR decoding hinders the fast generation of a lengthy response in practice. 
Speculative decoding~\cite{leviathan2023fast,zhou2023distillspec,liu2023online} and Medusa~\citep{cai2024medusa} are existing remediations to such an issue, but the former suffers from the difficulties in finding a suitable draft model and managing both models in a single system, and the latter causes significant increases in model size and architecture.  




In comparison, Jacobi decoding has shown the capacity to reduce the inference cost of LLMs without extra model components~\cite{santilli2023accelerating} and is therefore more applicable. 
Concretely, supposing $f(y_i, \vy_{<i}, \vx):= y_i-{\mathrm{arg\,max}_y}\ p(y | \vy_{<i}, \vx)$, Jacobi decoding re-frames the LLM inference process in \cref{eq:ar_decoding} as solving a system of nonlinear equations w.r.t. $y_i$:
\begin{equation}
f(y_i, \vy_{<i}, \vx) = 0 \;\, \text{for}\,\, i = 1,\dots,n. 
\end{equation}
It can be solved in parallel using the Jacobi fix-point iteration method~\citep{ortega2000iterative}, starting from a randomly initialized $n$-token sequence $\vy^{(0)}=\{y_{1}^{(0)},  \ldots, y_{n}^{(0)} \}$ and iteratively updating it by the following rule:
\begin{equation}
\label{eq:jacobi_decoding}
\begin{aligned}
\begin{cases}
y_{1}^{(j+1)} &= \underset{y}{\mathrm{arg\,max}}\ p(y | \vx) \\
y_{2}^{(j+1)} &= \underset{y}{\mathrm{arg\,max}}\ p(y | \vy_{1}^{(j)}, \vx) \\
& \vdots \\
y_{n}^{(j+1)} &= \underset{y}{\mathrm{arg\,max}}\ p(y | \vy_{<n}^{(j)}, \vx).
\end{cases}
\end{aligned}
\end{equation}
Notably, for LLM, the above $n$ maximization problems can be solved in parallel by using a causal attention mask, i.e., only one forward pass of the LLM is required to obtain $\vy^{(j+1)}$ based on $\vy^{(j)}$. 
The iteration exits at some $k$ such that $\vy^{(k)}=\vy^{(k-1)}$ and we define $\vy^*:=\vy^{(k)}$ as the fixed point. 
Let $\mathcal{J} := \{\vy^{(1)}, \dots, \vy^{(k)}\}$ denote the Jacobi trajectory. 
It can be proven that $\vy^*$ is identical to AR decoding under greedy strategy~\citep{song2021accelerating}. 
The acceleration effect of Jacobi decoding primarily stems from the fact that each forward pass of the LLM could potentially generate more than one fixed token within the $n$-token sequence, so the number of queries to the LLM could be smaller than that of AR decoding, i.e., $k \leq n$. 

Generally, for a prefix $\vx$ of length $n_x$, each forward pass in Jacobi decoding deals with a longer sequence of length $n_x + n$, demanding more FLOPs than AR decoding that deals with a shorter sequence length at $n_x + i$, $1 \leq i \leq n$. 
Yet, the added overhead can be minimal when $n_x$ is large or $n$ is small. 
Besides, we can integrate the KV cache mechanism~\cite{pope2023efficiently} into Jacobi decoding to further reduce the additional overhead, as detailed below. 

\textbf{Jacobi Decoding with KV Cache.} The sequential nature of LLMs ensures that each token generation is dependent only on preceding tokens. 
Namely, we have an increasing number of fixed tokens, which are correctly aligned with the AR generations. 
We don't need to iteratively update them and recompute their keys and values for computing attention in subsequent iterations thanks to the KV cache technique. 
So, we 1) progressively reduce the length of the iteration state by at least one token and 2) save the KV cache of fixed tokens along with the decoding procedure. 
We elaborate on this in \cref{alg:accelerated_jacobi}.

\subsection{Consistency Large Language Models (CLLMs)}
\label{sec:cllm}
Despite the promise, the speedup effect of Jacobi decoding for vanilla LLMs is minimal in practice~\cite{santilli2023accelerating,fu2024break}.
The reason is that AR-trained LLMs can usually generate only one correct token in each Jacobi iteration as such models can rarely yield a correct token when there are incorrect preceding tokens. 
To address this, we propose to adapt pre-trained LLMs to consistently map any point $\vy$ on the Jacobi trajectory $\mathcal{J}$ to the fixed point $\vy^*$. 
Surprisingly, such an objective is analogous to that of consistency models~\citep{songconsistency,song2023improved}, a leading acceleration approach for diffusion models~\cite{ho2020denoising,song2021scorebased}.

This section first delineates our data preparation procedure for tuning CLLM and then elaborates on the training procedure of CLLM. 
Lastly, we discuss some possible sources of the reason for CLLMs' acceleration.



\subsubsection{Jacobi Trajectory Collection} 
\label{sec:cllm dataset}

Let $p$ denote the target LLM we aim to adapt. Let $q_\theta(\cdot| \vx)$ denote the CLLM with parameters $\theta$ initialized with those of $p$. 
To realize the aforementioned adaptation, we collect a set of Jacobi trajectories by running the Jacobi decoding algorithm with the target LLM $p$ on prompts from a certain domain of interest, forming an original training set $\mathcal{D}$.
We summarize the algorithm for dataset generation in \cref{alg:dataset_generation}. 
Note that to generate a lengthy response $\vl$ of $N$ ($N \gg n$) tokens, we can sequentially perform Jacobi decoding for every truncation of $n$ tokens to avoid slow model evaluation on lengthy input. 
Consequently, $\vl$ amounts to the concatenation of a set of consecutive fixed points. 


\textbf{Data augmentation.} In a typical Jacobi iteration process, the correct tokens often appear one after another, and $n$-token sequences usually exhibit a ``correct, correct, wrong, wrong, wrong'' pattern. 
In comparison, patterns like ``correct, correct, wrong, correct, wrong'' can be rare. 
To enhance the learning and generalization capabilities of CLLMs, we augment the dataset $\mathcal{D}$ by randomly correcting erroneously predicted tokens within the samples.

\begin{algorithm}[t]
   \caption{Generate dataset to train a CLLM}
   \label{alg:dataset_generation}
\begin{algorithmic}
\small
   \STATE {\bfseries Input:} prompt set $\mathcal{O}$, n-token sequence size $n$, max new tokens $N$, target LLM $p$
   \REPEAT
   \STATE Sample prompt $\vx$ from origin dataset $\mathcal{O}$.
   \WHILE{$<$EOS$>$ is not generated and length generated $<N$}
    \STATE $\mathcal{J}=\{\vy^{(0)}, \ldots, \vy^{*}\} \leftarrow$ 
 Jacobi Decoding($p, \vx$)
    \STATE $\vx \leftarrow$ cat($\vx$, $\vy^{*}$)
   \IF{use data augmentation}
   \FORALL{$\vy \in \mathcal{J}$}
   \STATE Augment $\vy$ with false tokens corrected randomly
   \ENDFOR
   \ENDIF
   \STATE Append $\vx$ and $\mathcal{J}$ to Training Dataset $\mathcal{D}$
   \ENDWHILE
   \UNTIL{all prompts in origin dataset $\mathcal{O}$ are used}
\end{algorithmic}
\end{algorithm}

\textbf{Data post-processing.} 
Since the target LLM itself can make errors for some prompts, it often leads to low-quality generations in the Jacobi trajectories. We find training a CLLM with $n$-token sequences with token-level ~\citep{holtzman2019curious} or sentence-level repetitions~\citep{polivsenska2015sentence} often results in to repetitive content generation and noticeably degrades performance.
Recognizing the significance of high-quality datasets for training LLMs~\cite{zhou2023lima}, we perform post-processing to eliminate the low-quality samples from our training dataset $\mathcal{D}$ based on a rule-based detector. 

\subsubsection{Training}
\label{sec:cllm distillation}

We jointly optimize two losses for tuning CLLMs, one guaranteeing the prediction of multiple tokens at once and the other avoiding the CLLM from deviating from the target LLM so as to maintain generation quality. 

\textbf{Consistency Loss.} 
For a prompt $\vx$ with the Jacobi trajectory $\mathcal{J}$, let $\vy$ and $\vy^*$ denote a random state on the trajectory and the fixed point respectively. 
We can directly push CLLM to output $\vy^*$ with $\vy$ as the input by minimizing the following loss:
\begin{equation}
    \label{eq:global_loss}
    \begin{aligned}
    \mathcal{L}_{\text{GC}} &= \E_{ (\vx, \mathcal{J}) \sim \mathcal{D}, \vy \sim \mathcal{J}} \Big[ \\ &\quad\quad\quad\quad \sum_{i=1}^n D\left( q_{\theta^{-}}(\cdot|\vy_{<i}^{*}, \vx) || q_{\theta}(\cdot|\vy_{<i}, \vx) \right) \Big]
    \end{aligned}
\end{equation}
where $\theta^{-} = \text{stopgrad}(\theta)$ and we abuse notations to represent uniform sampling from the dataset. $D(\cdot||\cdot)$ denotes the distance between two distributions, with forward KL, reverse KL, and their mixture (\textit{i.e.}, the Jensen-Shannon divergence) as popular examples~\citep{agarwal2023gkd}. We primarily experiment with the forward KL. 

Alternatively, we can also achieve the goal that CLLM consistently maps all intermediate states to the fixed point with a local consistency (LC) loss following CMs~\cite{songconsistency}, where the adjacent states $(\vy^{(j)}, \vy^{(j+1)}$ in the Jacobi trajectory $\mathcal{J}$ are demanded to yield the same outputs:
\begin{equation}
    \label{eq:local_loss}
    \begin{aligned}
    \mathcal{L}_{\text{LC}} &= \mathbb{E}_{(\vx, \mathcal{J}) \sim \mathcal{D}, (\vy^{(j)}, \vy^{(j+1)}) \sim \mathcal{J}}\Big[ \\
    &\quad\quad \sum_{i=1}^nD\left(q_{\theta^{-}}(\cdot|\vy_{<i}^{(j+1)}, \vx)|| 
     q_{\theta}(\cdot|\vy_{<i}^{(j)}, \vx) \right)
     \Big].
    \end{aligned}
\end{equation}
We compare $\mathcal{L}_{\text{GC}}$ and $\mathcal{L}_{\text{LC}}$ empirically in \cref{tab:loss_design}, where the results show that the global consistency loss is more efficacious to train CLLMs. This is probably attributed to that $\mathcal{L}_{\text{LC}}$ only implicitly aims at mapping from any point consistently to the fixed point by minimizing the distance between consecutive points. However, there is still a gap between $\mathcal{L}_{\text{LC}}$ and the goal of predicting multiple tokens at once, because there is typically only one more correct token in $\vy^{(j+1)}$ than $\vy^{(j)}$ in the collected Jacobi trajectory.

\textbf{AR Loss.} To avoid deviating from the distribution of the target LLM, we incorporate the traditional AR loss based on the generation $\vl$ of the target LLM $p$:
\begin{equation}
    \label{eq:ar_loss}
    \mathcal{L}_{\text{AR}} = \E_{(\vx, \vl) \sim \mathcal{D}} \Big[ -\sum_{i=1}^N \log q_{\theta}(l_i|\vl_{<i}, \vx)\Big].
\end{equation}
This term contributes to maintaining generation quality substantially (see \cref{tab:loss_design}). 

Consequently, the total loss for training a CLLM is:
\begin{equation}
    \mathcal{L}(\theta) = \mathcal{L}_{\text{consistency}} + w\mathcal{L}_{\text{AR}}
\end{equation}
where $\omega$ represents a weighting coefficient, $\mathcal{L}_{\text{consistency}}$ can be either $\mathcal{L}_{\text{GC}}$ or $\mathcal{L}_{\text{LC}}$ and we adopt $\mathcal{L}_{\text{GC}}$ in our experiments.

The training procedure is detailed in \cref{alg:training_algorithm}.

\begin{algorithm}[t]
    \caption{Training algorithm for a CLLM}
    \label{alg:training_algorithm}
\begin{algorithmic}
\small
   \STATE {\bfseries Input:} Jacobi trajectory dataset $\mathcal{D}$, n-token sequence size $n$, the weight factor $\omega$, CLLM $q_\theta(\cdot |\vx)$
   \REPEAT
   \STATE Sample prompt $\vx$, Jacobi trajectory $\mathcal{J}$, and full response $\vl$ from $\mathcal{D}$
   \STATE Calculate $\mathcal{L}_{\text{AR}}$ using \cref{eq:ar_loss}
   \STATE Sample $\vy$ from $\mathcal{J}$
   \STATE Calculate $\mathcal{L}_{\text{consistency}}$ using \cref{eq:global_loss} or \cref{eq:local_loss}
    \STATE Calculate $\mathcal{L}(\theta)$ and update the parameters $\theta$
   \UNTIL{convergence}
\end{algorithmic}
\end{algorithm}

\subsection{Acceleration Mechanisms in CLLMs}
\label{sec:insights}

Next, we compare the Jacobi trajectory of the target LLM and CLLM in \cref{fig:trajectory_compare} to chase an in-depth understanding of acceleration mechanisms in CLLMs. 

As shown in the left side of \cref{fig:trajectory_compare}, target LLMs typically generate only one correct token in one iteration. In contrast, we identify \emph{fast forwarding} phenomenon where multiple consecutive tokens are correctly predicted in a single forward pass in CLLMs. The average fast forward count per forward pass in CLLMs ranges from $2$ to $6$ tokens as evaluated in~\cref{tab:profiling}. Moreover, tokens correctly generated in advance (\textit{e.g.} ``country'' and ``H'' in point $5$ and $6$ in the left side of \cref{fig:trajectory_compare}), are often replaced inaccurately in subsequent iterations in target LLMs. Unlike the pre-trained models, CLLMs exhibit the capability of predicting correct tokens preemptively, even with preceding incorrect tokens, while ensuring the tokens remain unchanged. We term such tokens as \emph{stationary tokens}, whose existance allow simultaneous extension of discontinuous correct tokens within the $n$-token sequence. Both phenomena contribute to the fast convergence in Jacobi decoding of CLLMs, thereby leading to a considerable generation speedup.

We observe that CLLMs acquire a crucial linguistic concept through training – \textbf{collocations}: a series of words or terms that co-occur more frequently than one would expect by random chance~\citep{smadja1991n}. Language is not solely composed of isolated words but also relies heavily on specific word pairings. Examples of collocations are abundant in both natural and coding languages. They include verb + preposition combinations (\textit{e.g.}, ``talk to'', ``remind ... of ...''), verb + noun structures (\textit{e.g.}, ``make a decision'', ``catch a cold''), and many more domain-specific syntactical structures (\textit{e.g.}, ``SELECT ... FROM ...'', ``if ... else'' for programming). The consistency generation objective allows CLLMs to infer such structures from any point in the Jacobi trajectory, encouraging CLLMs to acquire proficiency in numerous collocations and thereby predict multiple words simultaneously to minimize iteration steps. 

Notably, lookahead decoding~\citep{fu2024break} collects n-grams generated from previous Jacobi iterations as candidate tokens and verifies them in the next iteration to accelerate decoding. CLLMs can also be combined with lookahead decoding and achieve extra speedup (see \cref{tab:benchmarking_llama2} and \cref{tab:benchmarking_deepseek}) because collocations learned in CLLMs improve the quality of n-grams and thus increase the acceptance rate.



\section{Experiments}


\subsection{Evaluations} 

\begin{table}[t]
\centering
\caption{
Comparison of CLLMs with other baselines including speculative decoding using distilled draft model, Medusa, and fine-tuned model using LLaMA2-7B as the backbone model. Performance and inference speed are evaluated with applicable generation techniques. To quantify speed improvements, we measure speedup as the ratio of the wall-clock speed to the baseline AR decoding speed for each model. Results are measured with a batch size of 1. 
}
\vskip 0.1in
\small
\label{tab:benchmarking_llama2}
\begin{tabular}{ccccc}
\toprule
\text{Methods}  & \text{Speed (tokens/s)} & \text{Speedup} & \text{Metric}  & \text{Size}      \\ 
\midrule
\multicolumn{5}{c}{ \text{GSM8K} } \\
\midrule
\multicolumn{5}{c}{ \text{Fine-tuned LLaMA2-7B}~\cite{abel} } \\
\text{+ AR}  & 43.5  & 1.0$\times$ &  59.1 &     \multirow{3}{*}{6.7B}      \\
\text{+ Jacobi} &  45.7 & 1.1$\times$ & 59.1 &     \\
\text{+ lookahead} &  74.8 & 1.7$\times$ & 59.1 &      \\
\midrule
\multicolumn{5}{c}{ \textbf{CLLM-LLaMA2-7B} } \\
\text{+ AR}  & 43.5  & 1.0$\times$ & 56.4 &    \multirow{3}{*}{6.7B}    \\
\text{+ Jacobi} &  132.4  & \textbf{3.0$\times$} & 56.4 &         \\
\text{+ lookahead} & 125.2  & 2.9$\times$ & 56.4 &      \\
\midrule
\multicolumn{5}{c}{ \text{Medusa-2 + LLaMA2-7B} } \\
\text{+ typical}  & 70.2  & 1.6$\times$ &  51.3  &   8.3B       \\
\midrule
\multicolumn{5}{c}{ \text{Fine-tuned LLaMA2-7B + distilled LLaMA-160m} } \\
\text{+ speculative} &  73.8 & 1.7$\times$ & 59.1 & 6.8B   \\
\midrule
\midrule
\multicolumn{5}{c}{ $\text{ShareGPT (MT-Bench)}$ } \\
\midrule
\multicolumn{5}{c}{ \text{Fine-tuned LLaMA2-7B} } \\
\text{+ AR}  & 37.6 & 1.0$\times$ & 6.5 & \multirow{3}{*}{6.7B} \\
 \text{+ Jacobi} & 39.9 & 1.1$\times$ & 6.5  & \\ 
  \text{+ lookahead} & 60.8 & 1.6$\times$ & 6.5 &  \\
\midrule
\multicolumn{5}{c}{ \textbf{CLLM-LLaMA2-7B} } \\
\text{+ AR}  & 36.7  & 1.0$\times$ & 6.4 & \multirow{3}{*}{6.7B} \\ 
\text{+ Jacobi} & 88.4 & 2.4$\times$ & 6.4 &  \\ 
\text{+ lookahead} & 95.0  & \textbf{2.5$\times$} & 6.4 &  \\
\midrule
\multicolumn{5}{c}{ \text{Medusa-2 + LLaMA2-7B} } \\
\text{+ typical}  & 102.5  & 2.7$\times$ & 6.4 & 8.3B  \\
\midrule
\multicolumn{5}{c}{ \text{Fine-tuned LLaMA2-7B + distilled LLaMA-160m} } \\
\text{+ speculative} &  51.3 & 1.4$\times$ & 6.5 & 6.8B \\ 
\bottomrule
\end{tabular}
\vskip -0.1in
\end{table}

\begin{table}[t]
\centering
\caption{
Comparison of CLLMs with other baselines using Deepseek-Coder-7B-Instruct as the backbone model.
}
\vskip 0.1in
\small
\label{tab:benchmarking_deepseek}
\begin{tabular}{ccccc}
\toprule
\text{Methods}  & \text{Speed (tokens/s)} & \text{Speedup} & \text{Metric}  & \text{Size}      \\ 
\midrule
\multicolumn{5}{c}{ \text{Spider} } \\
\midrule
\multicolumn{5}{c}{ \text{Fine-tuned Deepseek-7B} } \\
\text{+ AR}  & 38.0  & 1.0$\times$ &  70.0   &   \multirow{3}{*}{6.7B}  \\
\text{+ Jacobi} & 39.5  & 1.0$\times$ & 70.0  &         \\
\text{+ lookahead} & 55.3  & 1.5$\times$ & 70.0 &       \\ 
\midrule
\multicolumn{5}{c}{ \textbf{CLLM-Deepseek-7B} }  \\
\text{+ AR}  &  38.0  & 1.0$\times$ & 69.3 &       \multirow{3}{*}{6.7B}  \\
\text{+ Jacobi} &  127.4  & 3.4$\times$ & 69.3 &         \\
\text{+ lookahead} &  135.2 & \textbf{3.6$\times$} & 69.3  &      \\
\midrule
\multicolumn{5}{c}{ \text{Medusa-2 + Deepseek-7B} } \\
\text{+ typical}  & 104.2   & 2.7$\times$ & 66.4 &      8.3B    \\
\midrule
\multicolumn{5}{c}{ \text{Fine-tuned Deepseek-7B + distilled LLaMA-160m} } \\
\text{+ speculative} &  66.8 & 1.8$\times$ & 70.0 &  6.8B  \\
\midrule
\midrule
\multicolumn{5}{c}{ \text{Code-Search-Net Python} } \\
\midrule
\multicolumn{5}{c}{ \text{Fine-tuned Deepseek-7B} } \\
 \text{+ AR}  & 40.1  & 1.0$\times$ & 60.4 & \multirow{3}{*}{6.7B} \\
 \text{+ Jacobi} & 43.2  & 1.1$\times$ & 60.4 & \\ 
\text{+ lookahead} & 68.0  & $1.7\times $ & 60.0&   \\
\midrule
\multicolumn{5}{c}{ \textbf{CLLM-Deepseek-7B} } \\
\text{+ AR}  &  38.5  & 1.0$\times$ & 59.2 & \multirow{3}{*}{6.7B} \\
\text{+ Jacobi} & 102.1  & 2.5$\times$ & 59.2 & \\ 
\text{+ lookahead} & 115.7  & \textbf{2.9$\times$} & 59.2 &  \\
\midrule
\multicolumn{5}{c}{ \text{Medusa-2 + Deepseek-7B} } \\
\text{+ typical}  &  128.0 & 3.2$\times$ & 48.3 & 8.3B \\  
\midrule
\multicolumn{5}{c}{ \text{Fine-tuned Deepseek-7B + distilled LLaMA-160m} } \\
 \text{+ speculative} &  59.3 & 1.5$\times$ & 60.4 & 6.8B \\ 
 \bottomrule

\end{tabular}
\vskip -0.1in
\end{table}

\textbf{Benchmarks and Setup.} We evaluate performance across three domain-specific tasks, including text-to-SQL (Spider)~\citep{yu2018spider}, Python code generation (Code-search-Python)~\citep{husain2019codesearchnet} and graduate school math (GSM8k)~\citep{cobbe2021gsm8k}. To test CLLMs generalizability on open-domain conversational interactions and instruction-following scenarios, we also train CLLMs on ShareGPT\footnote{\href{http://www.sharegpt.com.}{http://www.sharegpt.com}.} data and perform evaluation on the MT-bench~\citep{zheng2023judging}. The performance metrics are the greedy answers' problem solve rate (test@1) on GSM8K, MT-bench score, execution accuracy on Spider, as well as and strict accuracy (pass@1) on Human-Eval. Additionally, we also run evaluations of CLLMs' language modeling capability on raw-WikiText2~\citep{merity2016pointer} and PTB~\citep{pan2020meta}. 

Reported experiments were conducted using either pre-trained coder LLM, Deepseek-coder-7B-instruct~\citep{bi2024deepseek} or LLaMA-2-7B~\citep{touvron2023llama, touvron2023llama2} depending on the task. Both training and evaluation are carried out on servers equipped with 8 NVIDIA A100 40GB GPUs and 128 AMD EPYC 7742 64-core processors.

\textbf{Baselines.} In this section, we compare CLLMs with a range of alternative models that employ various strategies to speed up the inference process. This includes Medusa~\citep{cai2024medusa}, which modifies the underlying architecture, and approaches utilizing distilled draft models for speculative decoding~\citep{zhou2023distillspec, liu2023online}. Alongside these, we also consider fine-tuned baseline models for a comprehensive comparison. Our evaluation tests each model under different decoding paradigms the model is compatible with to thoroughly assess their inference quality and speed. The decoding algorithms include vanilla AR decoding, Jacobi decoding~\citep{song2021accelerating}, speculative decoding~\citep{leviathan2023fast}, and lookahead decoding~\citep{fu2024break}.

\textbf{Results.} To evaluate the performance and inference speedup of CLLMs across various tasks, we conduct an extensive comparison with the SOTA baselines on the three domain-specific tasks and the open-domain MT-bench. 

Table~\ref{tab:benchmarking_llama2} and Table~\ref{tab:benchmarking_deepseek} compare CLLMs against fine-tuned baseline models across three different generation modes: AR decoding, Jacobi decoding, lookahead decoding, and the stronger speculative decoding baseline using a distilled draft model. In both Jacobi and lookahead decoding, CLLMs consistently surpass the baselines. Notably, on the Spider dataset, CLLMs achieve a 3.4$\times$ speedup with negligible performance loss using Jacobi decoding. When benchmarked against other SOTA methods for efficient LLM inference, particularly those necessitating training, CLLMs exhibit the ability of fast consistency generation while maintaining lower memory and computational demands with lowest memory consumption in comparison with Medusa and speculative decoding. In these cases, we can still see CLLMs consistently outperform speculative decoding with distilled draft model and achieve better accuracy with comparable and even better inference speedup on datasets like Spider and GSM8K, where collocations are more common. CLLMs can also seamlessly integrate with
lookahead decoding and more speedup is gained compared to lookahead decoding applied in fine-tuned LLMs.

We highlight CLLMs' advantage over speculative decoding with distilled draft models and Medusa is its high adaptability. This is because CLLMs' are models tailored for Jacobi decoding. Jacobi decoding requires no modification to the original models. In the contrary, both speculative decoding and Meudsa require either auxiliary components like LM head, tree-based attention mask, or draft model, which usually come with the cost of searching for the optimal configuration. This is further summarized in Table~\ref{tab:baseline_comparison}.

Moreover, the language modeling results in Table~\ref{tab:benchmarking_language_modeling} show CLLMs are able to maintain a low perplexity while rendering at least $2\times$ speedup, suggesting CLLMs' potential to be trained as pre-trained LLM with higher inference efficiency.




\subsection{Acceleration Mechanisms in CLLMs}

\begin{table*}[t]
\centering
\caption{
\textbf{Profiling results for fast-forwarded and stationary token counts in fine-tuned models and CLLMs.} The numbers are reported for each $n$-token sequence, with the best-performing model and an accompanying n-gram size. Fast-forwarded token count reported in the table includes the one token that will be predicted right even without fast-forwarding.
}
\vskip 0.1in
\small
\label{tab:profiling}
\begin{tabular}{c|c|c|c}
\toprule
\text{Models} & \text{$n$-token sequence length} & \text{Fast-forward token count} & \text{Stationary token count}     \\ 
\midrule
\multicolumn{4}{c}{ \text{Spider} } \\
\text{Fine-tuned Deepseek-coder-7B-instruct}  & 16 & 1.1  &  0.4  \\
\text{CLLM-Deepseek-coder-7B-instruct (size 16)} & 16 & 5.7  & 1.6 \\ 
\midrule
\multicolumn{4}{c}{ \text{Code-Search-Net Python} } \\
\text{Fine-tuned Deepseek-coder-7B-instruct} & 32 & 1.1  &  0.4  \\
\text{CLLM-Deepseek-coder-7B-instruct (size 32)} & 32 & 4.0  & 6.8 \\ 
\midrule
\multicolumn{4}{c}{ \text{GSM8K} } \\
\text{Fine-tuned LLaMA-2-7B}  & 16 & 1.1  &  0.1  \\
\text{CLLM-LLaMA-2-7B (size 16)} & 16 & 2.8  & 2.0 \\ 
\midrule
\multicolumn{4}{c}{ \text{ShareGPT} } \\
\text{Fine-tuned LLaMA-2-7B}  & 32 & 1.1  &  0.3  \\
\text{CLLM-LLaMA-2-7B (size 32)} & 32 & 2.2 & 4.8 \\ 
\bottomrule
\end{tabular}
\end{table*}

With insights provided in Section~\ref{sec:insights}, we investigate the fast-forwarding phenomenon and the emergence of stationary tokens in Jacobi decoding to provide further empirical evidences for our hypothesis. We compare fast-forwarded and stationary token counts in target LLMs and CLLMs across the four datasets in Table~\ref{tab:profiling}.

From the table, there is a consistent 2.0x to 6.8x improvement in both fast-forwarded token and stationary token counts across all four datasets. In particular, for domain-specific datasets, such improvement is much more significant than open-domain dataset profiled on MT-bench. The results align with the observations from Section~\ref{sec:insights}, where we see more distinctive collocations and easy syntactical structures like blank space, newline tokens, and repetitive special characters in specialized domains like coding as demonstrated in Figure~\ref{fig:trajectory_compare}, versus open-domain conversations in ShareGPT and MT-bench with a significantly more diverse set of collocations.


\subsection{Ablation Studies} 
In this section, we evaluate the impact of various hyperparameter selections on the performance of CLLMs. 

\textbf{Dataset sizes and generalizability.} In Section~\ref{sec:cllm dataset}, Jacobi trajectory datasets are collected to conduct training for efficient Jacobi decoding. \cref{tab:dataset_size_ablation} demonstrates larger Jacobi trajectory datasets bring more significant speedup, and the speedup gradually saturates as the dataset size scales.
Moreover, CLLMs trained with more data can perform well even at the $n$-token sequence lengths it's not trained on and introduce more deployment-time robustness. 


\begin{table*}[t]
\centering
\caption{Comparison the performance of CLLMs trained with different sizes of Jacobi trajectory datasets on ShareGPT.}
\label{tab:dataset_size_ablation}
\vspace{1ex}
\begin{sc}
\begin{tabular}{ccccccc}
\toprule
\multirow{2}{*}{Trajectory Count} & \multirow{2}{*}{MT-bench}  & \multicolumn{5}{c}{Inference speedup (varying lengths)}\\
\cmidrule{3-7}
 &   & 16 & 32 & 64 & 128 & 256 \\
\midrule
 20k   &  6.1 & 1.7$\times$ & 1.8$\times$ & 1.4$\times$ & 1.2$\times$ & 1.1$\times$ \\
 100k  &  6.4 & 2.5$\times$ & 2.4$\times$ & 2.1$\times$ & 2.0$\times$ & 1.5$\times$ \\
 500k  & 6.4 & 2.7$\times$ & 2.7$\times$ & 2.2$\times$ & 2.1$\times$ &  1.8$\times$ \\
\bottomrule
\end{tabular}
\end{sc}
\end{table*}

\begin{figure}[t]
\centering
\includegraphics[width=0.45\textwidth]{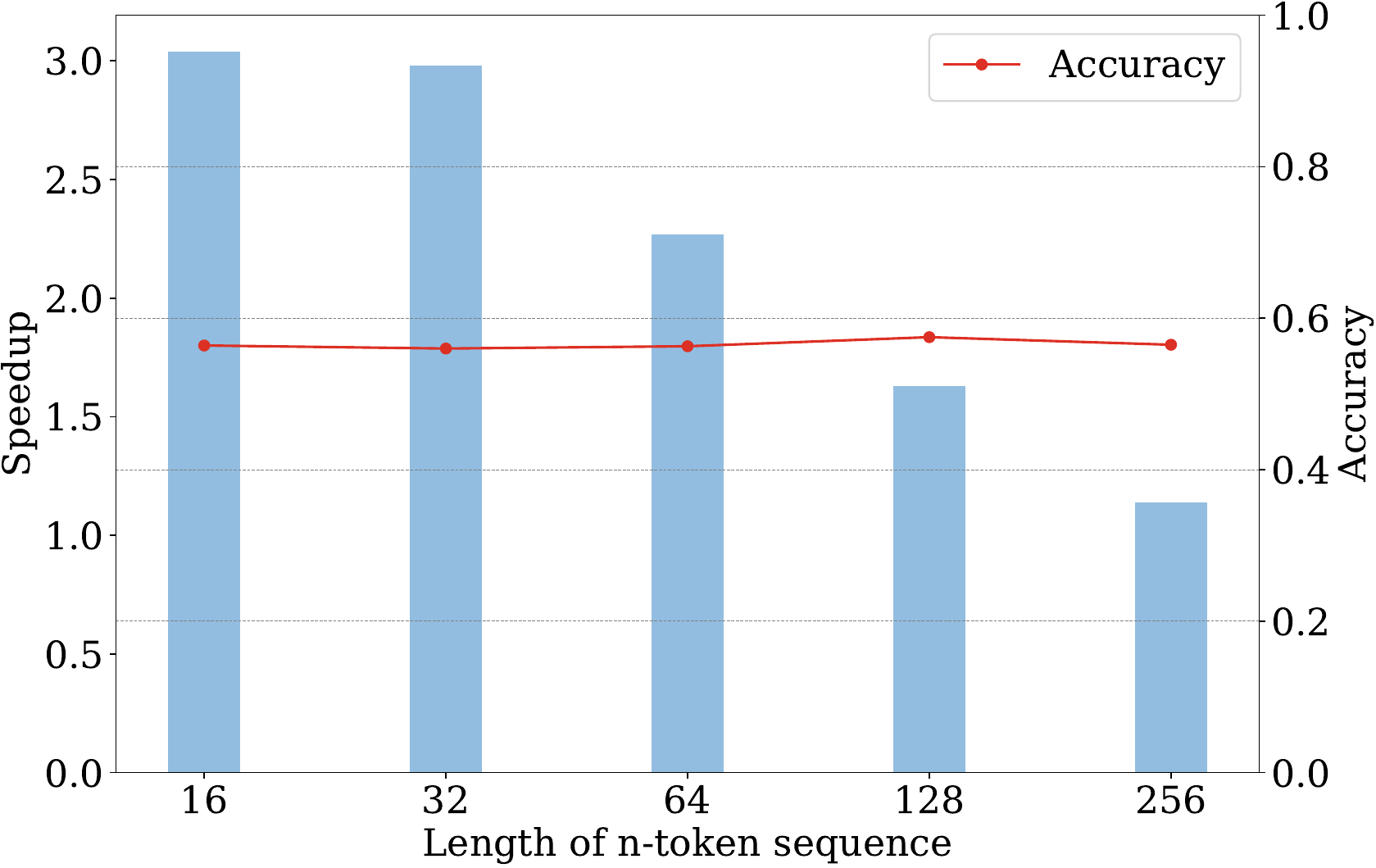} 
\caption{Accuracy and speedup of models trained with different n-token sequences lengths on GSM8K dataset. The sequence length for generation matches the training settings. Speedup is measured as the ratio of the wall-clock generation throughput when employing Jacobi decoding, and that of the baseline AR decoding. }
\label{fig:n-gram_ablation}
\vskip -0.1in
\end{figure}

\textbf{Different lengths of $n$-token sequence.} We investigate how different $n$-token sequence lengths in the Jacobi trajectory dataset affect CLLMs' performance on GSM8K. We employ varying lengths to generate the Jacobi dataset and train the CLLMs accordingly. Figure \ref{fig:n-gram_ablation} illustrates that CLLMs consistently maintain generation quality while the models are trained with different lengths. 
In practice, longer sequence lengths come at cost of increased computational overhead during inference. In Figure \ref{fig:n-gram_ablation}, significant degradation inference speed can thus be observed when the $n$-token sequence length exceeds 64. 

\label{sec:ablation}
\textbf{Loss design.} We adjust the ratio of consistency loss to autoregressive loss described in Section~\ref{sec:cllm distillation} and evaluate different loss ratios' performance on GSM8K. As illustrated in \cref{tab:loss_design}, increasing the emphasis on autoregressive loss does indeed enhance accuracy, though it slightly compromises the speedup gains. Additionally, we compare the efficacy of CLLMs using both consistency global loss and consistency local loss. \cref{tab:loss_design} demonstrates that the global loss is more efficacious in the training of CLLMs.

\subsection{Limitations and Discussion} 

\begin{table}[h]
\centering
\caption{
CLLMs' performance versus the fine-tuned baseline on language modeling tasks. 
}
\vskip 0.1in
\small
\label{tab:benchmarking_language_modeling}
\begin{tabular}{cccc}
\toprule
\text{Methods}  & \text{Speed (tokens/s)} & \text{Speedup}   & PPL $(\downarrow)$  \\ 
\midrule
\multicolumn{4}{c}{ \text{raw-WikTtext2} } \\
\midrule
\multicolumn{4}{c}{ \text{fine-tuned LLaMA2-7B} } \\
\text{+ AR}  & 41.2  & 1.0$\times$ &  8.0      \\
\text{+ Jacobi} &  36.9 &  1.0$\times$ & 8.0     \\
\text{+ lookahead} &  58.1 & 1.6$\times$ & 8.0    \\
\midrule
\multicolumn{4}{c}{ \text{CLLM-LLaMA2-7B} } \\
\text{+ AR}  & 40.1  & 1.0$\times$ &  9.5     \\
\text{+ Jacobi} &  83.2 &  2.1$\times$ & 9.5      \\
\text{+ lookahead} &  89.5 & 2.2$\times$ & 9.5       \\
\midrule
\midrule
\multicolumn{4}{c}{ \text{PTB} } \\
\midrule
\multicolumn{4}{c}{ \text{fine-tuned LLaMA2-7B} } \\
\text{+ AR}  & 43.8  & 1.0$\times$ &  15.6      \\
\text{+ Jacobi} &  41.8 &  1.0$\times$ & 15.6     \\
\text{+ lookahead} &  62.0 & 1.5$\times$ & 15.6      \\
\midrule
\multicolumn{4}{c}{ \text{CLLM-LLaMA2-7B} } \\
\text{+ AR}  & 43.6  & 1.0$\times$ &  15.3    \\
\text{+ Jacobi} &  98.1 &  2.3$\times$ & 15.3      \\
\text{+ lookahead} &  101.5 & 2.3$\times$ & 15.3    \\
\bottomrule
\end{tabular}
\end{table}

In our experiments, we observe that achieving significant speedup while maintaining good generation quality with a CLLM relies strongly on having a high-quality Jacobi trajectory dataset. Therefore, data cleaning is crucial, as discussed in Section~\ref{sec:cllm dataset}. Dataset size also plays a role as described in Section~\ref{sec:ablation} and shown in Table~\ref{tab:dataset_size_ablation}, although to a lesser extent. For instance, Jacobi trajectories generated with only 10\% of the Code-Search-Net Python dataset is able to yield a $2.9\times$ speedup as demonstrated in Table~\ref{tab:benchmarking_deepseek}. However, for open-domain datasets like ShareGPT, more data is necessary for improved efficiency. The computation cost for CLLMs training is moderate and discussed in Appendix~\ref{appendix:compute_cost}.

In our proposed method and experiments, we primarily use output sequences from the teacher ~\citep{kim2016sequence} to collect Jacobi trajectories and train a CLLM. This introduces some additional overhead in comparison with conventional model training. On-policy GKD proposed in \citet{agarwal2023gkd} suggests LLM distillation using a mixture of teacher and student samples or even student samples by themselves can yield high-performance models. One mitigation is therefore to use $n$-token sequences generated by the trained model itself as the training samples. This can remove the Jacobi trajectory collection overhead, making our proposed method potentially feasible for pre-training. 


Results from our language modeling experiments, as detailed in Table~\ref{tab:benchmarking_language_modeling}, demonstrate the robustness of the CLLM when trained on pre-training jobs with a notable speedup. By incorporating on-policy GKD, it is conceivable that a modified version of our proposed method could be employed for LLM pre-training. This modification would equip the pre-trained model with both a strong language modeling capability, as existing models possess, and a high generation speed when employing Jacobi decoding for inference. We leave the opportunities of adapting CLLMs to pre-trained jobs for future work. 

\begin{table}[t]
\caption{Comparison the performance of CLLMs trained with different loss design. All models are trained on GSM8K.}
\small
\label{tab:loss_design}
\begin{center}
\begin{sc}
\begin{tabular}{lcc}
\toprule
Loss & Speedup & Accuracy \\
\midrule
$\mathcal{L}_{\text{CTG}} + \mathcal{L}_{\text{AR}} $   & 3.2$\times$ & 51.3 \\
$\mathcal{L}_{\text{CTG}} + 10\cdot \mathcal{L}_{\text{AR}} $ & 3.0$\times$ & 56.4 \\
$\mathcal{L}_{\text{CTL}} + \mathcal{L}_{\text{AR}} $   & 2.8$\times$ & 55.2 \\
$\mathcal{L}_{\text{CTL}} + 10 \cdot \mathcal{L}_{\text{AR}} $   & 2.4$\times$ & 56.0 \\
\bottomrule
\end{tabular}
\end{sc}
\end{center}
\vskip -0.25in
\end{table}

\section{Conclusion}
In this work, we introduce CLLMs, a new family of LLMs that excel in efficient parallel decoding, designed to significantly enhance the efficiency of Jacobi decoding. Unlike other existing techniques for efficient LLM inference, which often require either additional architectural components~\citep{cai2024medusa, li2024eagle} or draft models~\citep{leviathan2023fast,zhou2023distillspec,liu2023online}, CLLMs are directly adapted from a target pre-trained LLM. This reduces the complexity associated with additional architecture designs or managing two different models in a single system. In addition, CLLMs can also be integrated seamlessly with other techniques for efficient LLM inference~\citep{dao2023flashattention,fu2024break,ainslie2023gqa} to achieve greater speedup. We have demonstrated the efficacy of CLLMs on both specific and open domains, revealing a significant improvement in generation speed while preserving generation quality.



\section*{Acknowledgments}
This work was supported by Key R\&D Program of Shandong Province, China (2023CXGC010112), NSF of China (Nos. 62306176, 62102257), National Key R\&D Program of China~(2022YFB4500200), Natural Science Foundation of Shanghai (No. 23ZR1428700), and CCF-Baichuan-Ebtech Foundation Model Fund.

\section*{Impact Statement}
This work presents a challenge in machine learning and proposes a solution, the potential negative consequences are not apparent. While it is theoretically possible for any technique to be misused, the likelihood of such misuse occurring at the current stage is low.   

\bibliography{Consistency_Decoding}
\bibliographystyle{icml2024}

\newpage
\appendix
\onecolumn

\section{Illustration of Consistency Loss Learning Objectives}
\label{appendix:learning_objective}

In our proposed method described in Section~\ref{sec:cllm}, we use Jacobi trajectories collected from a target model to train the model with a loss that encourages single-step convergence during Jacobi iterations. This is achieved with either choice of the two consistency loss:

\begin{list}{$\bullet$}{\leftmargin=1em \itemindent=0em}
    \item \textbf{Global consistency loss}: directly minimize the distance $D$ between any arbitrary point $\mathbf y$ on a Jacobi trajectory and the fixed point $\mathbf y^*$ in Equation~\ref{eq:global_loss}.
    \item \textbf{Local consistency loss}: minimize the distance $D$ between any arbitrary point $\mathbf y^{(j)}$ on a Jacobi trajectory with its adjacent state $\mathbf y^{(j+1)}$ in Equation~\ref{eq:local_loss}, which thereby also implicitly minimizes the distance between $\mathbf y^{(j+1)}$ and the fixed point $\mathbf y^*$.
\end{list}

\begin{figure*}[h]
\centering
\includegraphics[width=0.75\textwidth]
{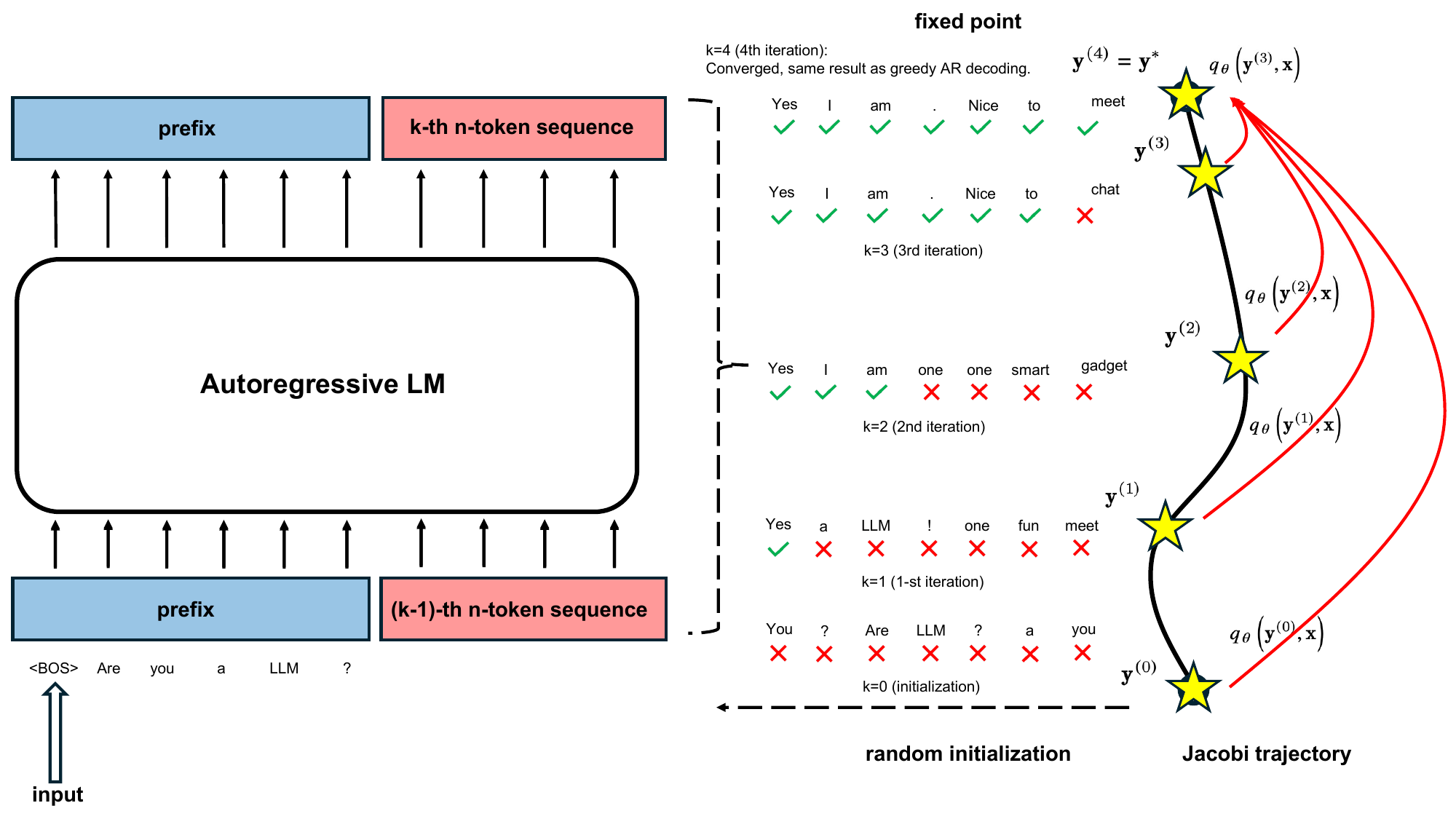} 
\caption{The image illustrates global consistency loss where we aim to directly learn a model $q_{\theta}$ that maps arbitrary $n$-token sequence $\mathbf y^{(0)}$, $\mathbf y^{(1)}$, etc.) to the fixed point $\mathbf y^{*}$.
}
\label{fig:objective_illustration_global}
\vskip -0.1in
\end{figure*}

\begin{figure*}[h]
\centering
\includegraphics[width=0.75\textwidth]
{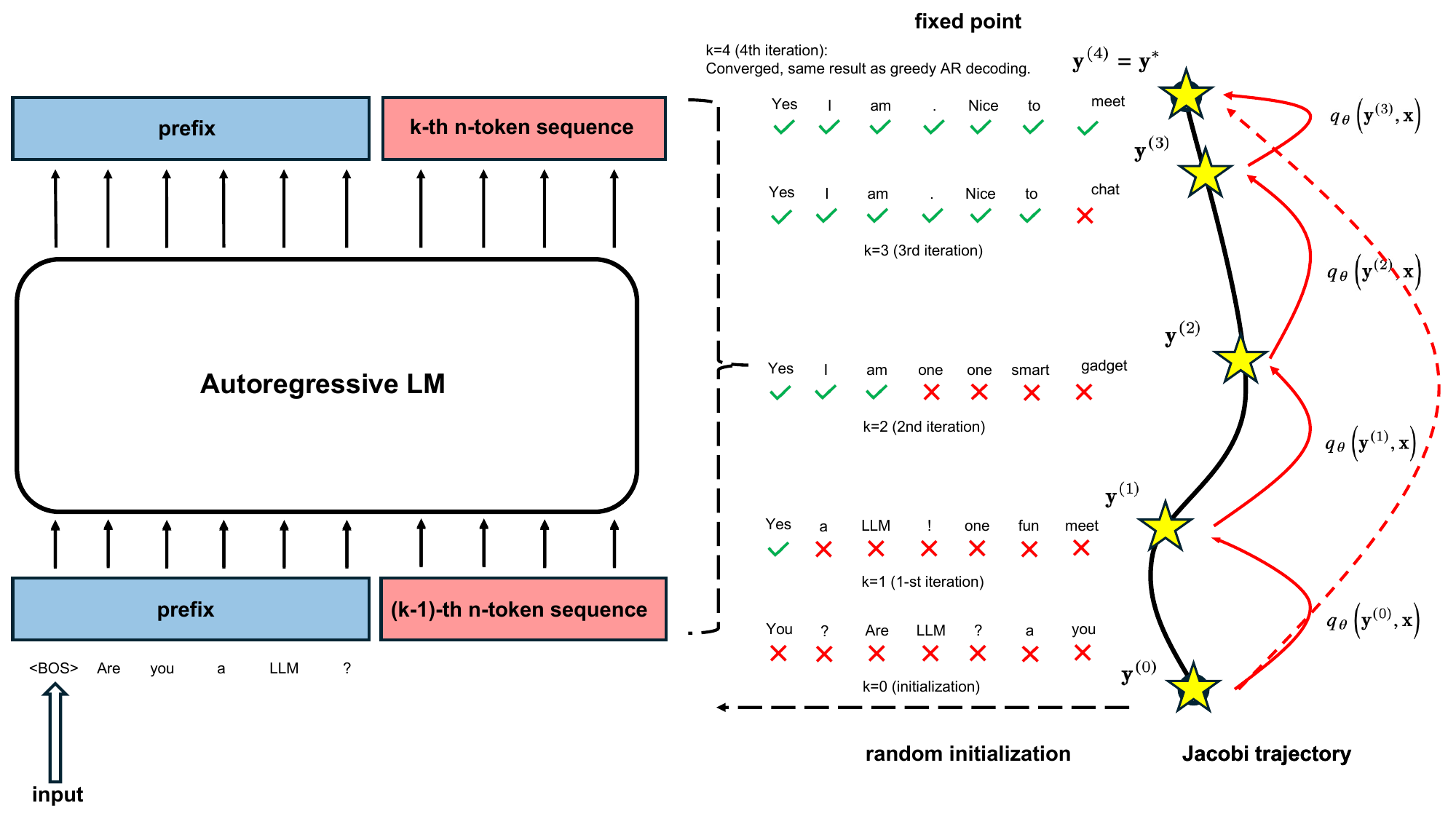} 
\caption{The image illustrates local consistency loss where we aim to learn a model $q_{\theta}$ that maps an arbitrary $n$-token sequence $\mathbf y^{(j)}$ to its next adjacent state, and implicitly mapping the point to the fixed point $\mathbf y^{*}$.
}
\label{fig:objective_illustration_local}
\vskip -0.1in
\end{figure*}

An illustration further depict the global consistency loss and the local consistency loss in Figure~\ref{fig:objective_illustration_global} and Figure~\ref{fig:objective_illustration_local}. 

\section{Comparison with Baseline Algorithms}
\label{appendix:baseline_comparison}

In this section, we present a comparative analysis of baseline algorithms for efficient LLM inference. Key features considered are listed below. Table~\ref{tab:baseline_comparison} underlines that CLLMs, our proposed method, stands out for its memory efficiency and adaptability, requiring no modifications to the existing model architecture while achieving up to $3.4\times$ inference speedup.
\begin{list}{$\bullet$}{\leftmargin=1em \itemindent=0em}
    \item \textbf{Lossless}: whether the method generates exactly the same output distribution as AR decoding does in the backbone model.
    \item \textbf{Training-free}: whether the method requires training.
    \item \textbf{Architecture-design-free}: whether the method requires modifications or adding auxiliary components to pre-trained LLMs (like extra MLP layers, LM heads~\citep{cai2024medusa}, autoregressive heads~\citep{li2024eagle}, etc.).
    \item \textbf{Attention-modification-free}: whether the methods require modifications to exisiting attention mechanism in transformers. For example, this includes tree token verification as appears in ~\citet{cai2024medusa}.
    \item \textbf{Extra-memory-free}: whether the method requires extra memory conmsumption in the system to accommodate speculative model or extra parameters.
    \item \textbf{Speedup}: Whether the method can effectively deliver inference speedup in practical use cases.
\end{list}

\definecolor{yesgreen}{rgb}{0.13, 0.55, 0}
\definecolor{nored}{rgb}{0.64, 0.12, 0.20}
\definecolor{maybeyellow}{rgb}{0.98, 0.70, 0.0}
\def\yes{\color{yesgreen}{yes}}
\def\no{\color{nored}{no}}
\def\maybe{\color{maybeyellow}{yes}}

\begin{table*}[h]
\centering
\small
\vskip -0.2in
\caption{
All speedups are relative to the vanilla AR. CLLMs has the best memory efficiency and adaptability as it requires no modifications to the model. $\text{\yes$^{*}$}$ refers to capability of achieving more than $3\times$ speedup on at least one of our benchmarks. Jacobi decoding doesn't always lead to a speedup as discussed in Section~\ref{sec:preliminary}, so we denote it with $\text{\maybe}$.
}
\vskip 0.1in
\begin{tabular}{ccccccc}
\toprule
Methods & Lossless & Training-free  &  Arch-design-free  & Attention-mod-free  & Extra-memory-free  & Speedup \\
\midrule
Vanilla AR  
      &    \yes &  \yes       &       \yes       &      \yes      &    \yes  &  \no \\
\midrule
Jacobi Decoding 
      &    \yes  & \yes      &        \yes       &    \yes      &     \yes     & \maybe   \\
\midrule
Speculative Decoding 
       &   \yes   & \yes     &       \yes       &     \yes       &      \no       &  \yes \\
\midrule
Lookahead Decoding 
      &   \yes   & \yes     &       \yes       &     \yes       &       \no       &  \yes \\
\midrule
SD with Distilled Student 
      &    \yes  & \no     &       \yes       &     \yes        &        \no        &  \yes \\
\midrule
Eagle
      &    \yes  &  \no     &       \no       &     \no        &      \no      &  \yes$^{*}$ \\
\midrule
Medusa 
      &    \no   & \no      &       \no       &    \no         &      \no     &  \yes$^{*}$ \\
\midrule
\textbf{CLLMs ( Ours )}
      &    \no    & \no     &       \yes       &   \yes        &     \yes     &  \yes$^{*}$ \\
\bottomrule
\end{tabular}
\label{tab:baseline_comparison}
\vspace{-5pt}
\end{table*}



\section{Pesudo Code for Jacobi Decoding with KV Cache} 
\label{appendix:kv_cache}

\begin{algorithm}[h]
   \caption{Jacobi Decoding with KV Cache}
   \label{alg:accelerated_jacobi}
\begin{algorithmic}[1]
\small
   \STATE {\bfseries Input:} prompt $\vx$, n-gram size $n$, past KV cache $\mathcal{K}$, LLM, Jacobi trajectory $\mathcal{J}$
   \STATE $\vy \gets \text{random tokens from}$ $\vx$
   \STATE $n_t \gets 0$ \COMMENT{Initialization of accurate length}
   \STATE $y_0,\mathcal{K} \gets$ LLM($\vx$) \COMMENT{Prefill phase: generate the first token}
   \STATE $\vz_{\text{next}} \gets $ cat$\left(y_0, \vy_{\geq 1}\right)$
   \REPEAT
      \STATE $\vz^{\text{current}} \gets \vz^{\text{next}}$        \STATE $\vz^{\text{next}}, \mathcal{K} \gets$ LLM($\vz^{\text{current}}, \mathcal{K}$)
      \STATE $i^{*} \gets \max \{ i \mid \vz^{\text{current}}_{<i} = \vz^{\text{next}}_{<i},\, i \in \{0, \ldots, $ $\text{len}(\vz^{\text{current}})-1\} \} $ \COMMENT{Fast-forwarded token count}
      \STATE $\vy_{n_t \leq i^{'} < n_t + i^*} \gets \vz^{\text{next}}_{<i^{*}}$ \COMMENT{$i^{'}$ denotes a dummy variable}
      \STATE $n_t \gets n_t + i^{*}$
      \STATE Append cat$\left(\vy_{<n_t}, \vz^{\text{next}}_{\geq i^{*}}\right)$ to $\mathcal{J}$
      \STATE Remove KV cache of false tokens from $\mathcal{K}$
      \STATE $\vz^{\text{next}} \gets \vz^{\text{next}}_{\geq i^{*}}$
   \UNTIL{$n_t = n$}
   \STATE {\bfseries Ouput:} $\mathcal{J}$ and $\vy$
\end{algorithmic}
\end{algorithm}

\section{Computation cost for CLLMs training}
\label{appendix:compute_cost}
\textbf{For the computation required for dataset generation,} the cost is low and it’s a one-time overhead. In the cases where the dataset size is large, for example for CodeSearchNet-Python, only $10\%$ of the dataset is required to generate Jacobi trajectories and the trained CLLMs obtain around 2.5× speedup on average. More details are shown in the table below. 

\begin{table*}[h]
\centering
\caption{
Computation required for dataset generation. The estimated generation time is based on sequential generation with batch size = 1. We can further reduce the generation time with serving systems like vLLM~\citep{kwon2023efficient} with batch size = 16 or more. We give an example of estimated training time with vLLM using batch size = 16 in the table as well. All time is estimated by a single A100 40G GPU*hours.
}
\vspace{2ex}
\begin{tabular}{lccc} 
\toprule
Dataset & 
$\sharp$ Generated tokens
& 
Estimated generation time
& 
Estimated generation time(vLLM) \\
\midrule Spider & $2 \mathrm{M}$  & 5 & $<1$ \\
\hline GSM8K & $10 \mathrm{M}$ & 14 & $\sim 1$ \\
\hline CodeSearchNet-Python & $100 \mathrm{M}$  & 100 &  8 \\
\hline ShareGPT & $200 \mathrm{M}$ & 120 & 10 \\
\bottomrule
\end{tabular}

\end{table*}

\textbf{For the computation required for consistency training,} we conclude the time and resources required for training a CLLM in the table below. For the percentage of pre-training cost, we estimate it by $\displaystyle \frac{ \sharp \text{tokens required for training a CLLM}}{\sharp \text{tokens required for pre-training}}$, where $\sharp$tokens required for pre-training is 1T for LLaMA-7B~\citep{touvron2023llama}.
\begin{table*}[h]
\centering
\caption{
Computation required for consistency training.
}
\begin{tabular}{lccc}
\toprule
Dataset & Training time & \% of pre-training cost & Training resources  \\
\midrule Spider & 2 hours & $<0.01 \%$ & 8 A100 40GB GPUs \\
\hline GSM8K & 12 hours & $\sim 0.01 \%$ & 8 A100 40GB GPUs \\
\hline CodeSearchNet-Python & 22 hours & $\sim 0.1 \%$ & 8 A100 40GB GPUs \\
\hline ShareGPT & 30 hours & $\sim 0.2$ \% & 8 A100 40GB GPUs \\
\bottomrule
\end{tabular}

\end{table*}

\end{document}